\documentclass[sigconf]{acmart}

\usepackage{balance}

\def\BibTeX{{\rm B\kern-.05em{\sc i\kern-.025em b}\kern-.08emT\kern-.1667em\lower.7ex\hbox{E}\kern-.125emX}}
    
\copyrightyear{2020}
\acmYear{2020}
\setcopyright{acmlicensed}
\acmConference[AIES '20]{Proceedings of the 2020 AAAI/ACM Conference on AI, Ethics, and Society}{February 7--8, 2020}{New York, NY, USA}
\acmBooktitle{Proceedings of the 2020 AAAI/ACM Conference on AI, Ethics, and Society (AIES '20), February 7--8, 2020, New York, NY, USA}
\acmPrice{15.00}
\acmDOI{10.1145/3375627.3375849}
\acmISBN{978-1-4503-7110-0/20/02}

\begin{document}

\fancyhead{}

\title{Deepfakes for Medical Video De-Identification: Privacy Protection and Diagnostic Information Preservation}

\author{Bingquan Zhu, Hao Fang, Yanan Sui, Luming Li}
\email{{zhubq17,fangh18}@mails.tsinghua.edu.cn, {ysui,lilm}@tsinghua.edu.cn}
\affiliation{%
  \institution{National Engineering Laboratory for Neuromodulation, Tsinghua University}
}


%
\renewcommand{\shortauthors}{Zhu et al.}

%
\begin{abstract}
Data sharing for medical research has been difficult as open-sourcing clinical data may violate patient privacy. Traditional methods for face de-identification wipe out facial information entirely, making it impossible to analyze facial behavior. Recent advancements on whole-body keypoints detection also rely on facial input to estimate body keypoints. Both facial and body keypoints are critical in some medical diagnoses, and keypoints invariability after de-identification is of great importance. Here, we propose a solution using deepfake technology, the face swapping technique. While this swapping method has been criticized for invading privacy and portraiture right, it could conversely protect privacy in medical video: patients' faces could be swapped to a proper target face and become unrecognizable. However, it remained an open question that to what extent the swapping de-identification method could affect the automatic detection of body keypoints. In this study, we apply deepfake technology to Parkinson's disease examination videos to de-identify subjects, and quantitatively show that: face-swapping as a de-identification approach is reliable, and it keeps the keypoints almost invariant, significantly better than traditional methods. This study proposes a pipeline for video de-identification and keypoint preservation, clearing up some ethical restrictions for medical data sharing. This work could make open-source high quality medical video datasets more feasible and promote future medical research that benefits our society.
\end{abstract}

%

%

%
\maketitle
\section{Introduction}
\noindent Deepfake technology has drawn a lot of attention recently\cite{fsgood}. It could swap faces between two persons in videos, and this method is being used to produce videos for entertainment. Public concerns are growing as deepfake technology may create inappropriate contents\cite{fsbad1} and infringe privacy rights\cite{fsbad2}. Its ability to challenge visual authenticity has a profound impact on our daily life. However, technologies could often do both evil and good, so as deepfakes. In this study, we demonstrate that deepfake technology can also be used to protect privacy in some important (e.g. medical) applications, working as a better performing face de-identification method.\\
\begin{figure}[t]
\centering
\includegraphics[width=0.95\columnwidth]{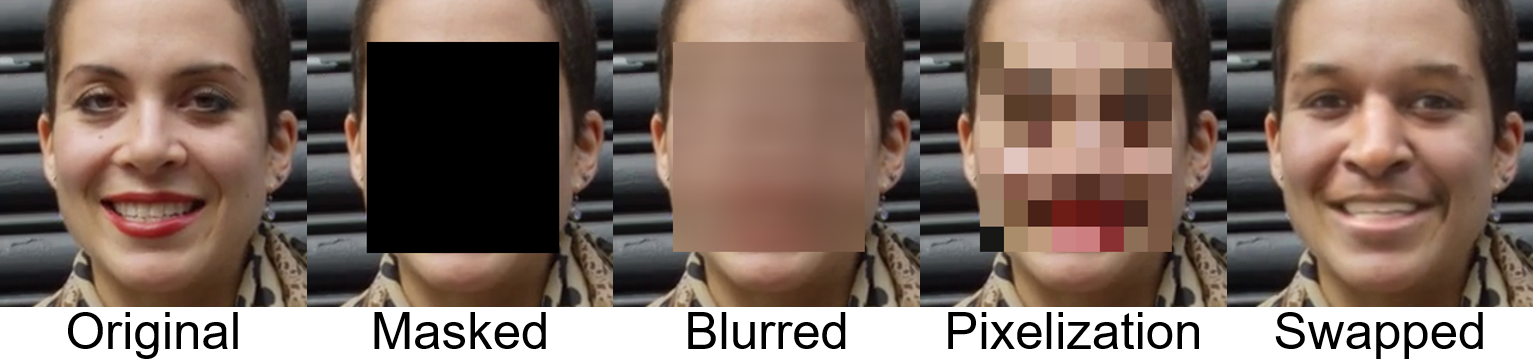} 
\caption{Images of various de-identification methods.}
\label{traditional}
\end{figure}
\indent Face de-identification has long been a topic in the study of computer vision\cite{deid1,deid2}. Traditional face de-identification methods include masking, blurring, and pixelization. Masking means covering a mask (usually a pure color box) to face region. Blurring and pixelization involve various smoothing or down-sampling operations to face regions (see Figure \ref{traditional}). The above methods directly cover face regions, completely eliminating facial data and any other objects moving around faces. The traditional methods result in unnatural images and lose all facial information, which are rather undesirable features. The emerging deepfake technology provides an all-in-one solution to facial information preservation and de-identification. Faceswap\cite{deepfakes} is a deepfake technology that takes in faces of two people and trains a deep neural network to swap their faces. If a subject's face is swapped to an open-source face, the generated face becomes unrecognizable to the original one. Meanwhile, the generated face still keeps original facial information, such as expressions and skin color. This approach is robust to different facial expressions, lighting conditions, and poses. It is also applicable to videos. \\
\indent Deepfake technology could be very suitable for some medical applications. Privacy and ethical issues greatly restrict medical image and video sharing when human faces are presented. This problem is very common in neurological diseases such as movement disorders. During video recording for diagnosis, it is often inevitable to record subjects' faces. Sometimes the faces of patients are clinical manifestation of movement disorder\cite{updrsvideo}. An alternative for sharing patient movement videos is to extract 3D keypoints and publish the keypoints data instead\cite{mpi3d-dataset,kit3d-dataset}. However, raw videos could provide far richer information than merely keypoints. In many cases, the raw videos of patients are way more reliable and valuable than keypoints extracted by algorithms. Also, state-of-the-art keypoint detection algorithms are still far from optimal and reliable tracking. For example, Parkinson's disease\cite{pd} is a nervous system disease that presents movement disorders, such as bradykinesia on face and tremor. Parkinson's disease manifestation is evaluated by the Unified Parkinson’s Disease Rating Scale (UPDRS)\cite{updrs}. The UPDRS examinations are often videotaped for further analysis. For instance, automated keypoint detection to assess the movement without markers is an emerging field. Lack of high quality video dataset hinders the research on movement disorders. As deepfakes becomes a new option for face de-identification, video data sharing could be more applicable than before.\\
\indent Independent of our work, an encoder-decoder network was recently proposed to maximally decorrelate identities of the original and converted videos for live face de-identification\cite{facebook}. This work mainly tries to cheat machine recognition methods, deliberately produces output images that look similar to the original subject in human perception, and shows no intention for medical information preservation. In our case, the face-swapped videos should both be de-identified by machine algorithms and humans.\\
\indent However, one thing has been unclear: are keypoints invariant during face swapping? Since state-of-the-art technologies of RGB image keypoint detection have full image receptive fields, it is likely that both detected facial keypoints and body keypoints change after facial modifications. Presumably, body keypoints detection with natural faces would be better than body keypoints detection with unnaturally covered facial areas. Major progress has been made in keypoint detection task with deep learning methods, for example, convolutional pose machine\cite{openpose-cpm}. Openpose\cite{openpose-multi-person,openpose-hand} is a widely used keypoint detection method that employs the convolutional pose machine. Research work using Openpose to extract pose keypoints for applications like medical analysis are emerging\cite{opgait,opgait2}. But it's known that the methods behave like black boxes and unpredictable changes in detected keypoints could occur even if a minor change is applied to the input. In medical applications, fine tracking of motion is required and it is important to minimize the changes that face-swapping process brings. Thus, it remained an open question that to what extent face-swapping could lead to a change in keypoint detection results.\\
\indent In this work, we apply face-swapping to medical videos for patient privacy protection and diagnostic human motion information preservation. We show that deepfake technology is a promising approach for de-identified invariable keypoint data sharing. We propose a pipeline to quantify the reliability of face de-identification and the invariability of keypoints. This approach makes high quality medical video sharing possible. Since the ethical issues are substantially alleviated in this way, the computer vision society and the medical society could better cooperate to conduct research on movement disorders.
\section{Method}
\begin{figure*}[t]
\centering
\includegraphics[width=0.95\textwidth]{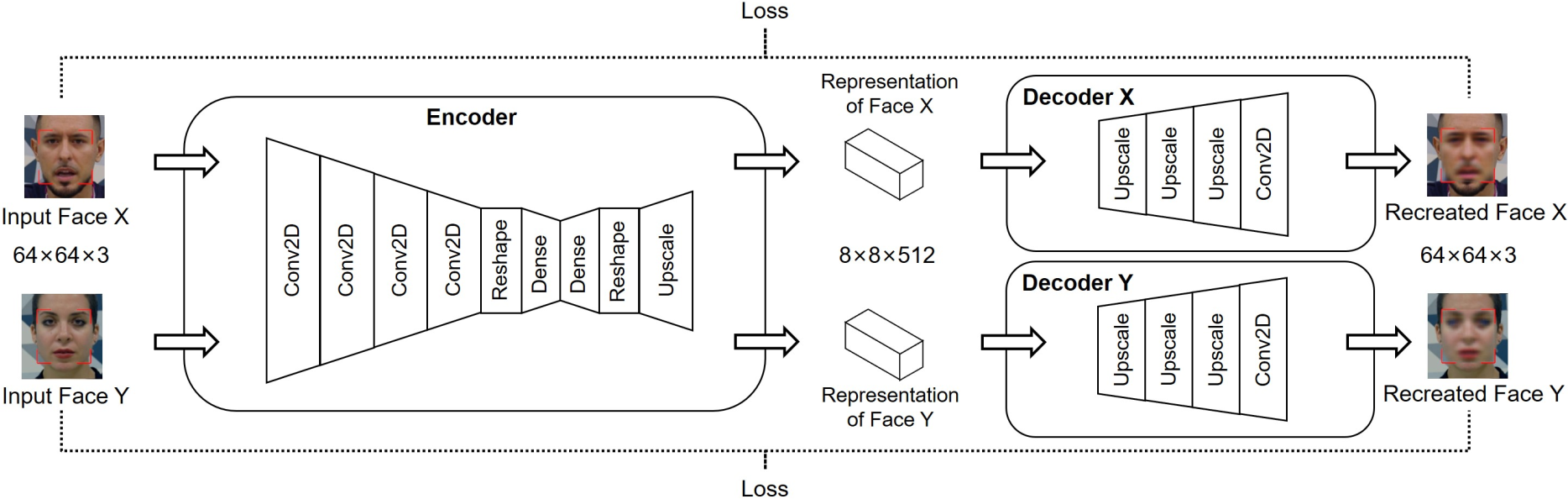} 
\caption{How Faceswap works. Here X and Y are open-source example subjects. The input face is processed by an encoder and becomes a representation vector. The vector is processed by a corresponding decoder and becomes a recreated face. During training, the encoder is shared between 2 subjects while the decoders are separated. Loss for backpropagation is calculated by the difference between input faces and recreated faces. During swapping, the model switches the decoder to generate swapped images.}
\label{ae-train}
\end{figure*}
There are four steps in our proposed pipeline:
\begin{enumerate}
\item We select videos of subject X (a patient) and subject Y (an open-source character) (original videos).
\item We train a Faceswap model of subjects X\&Y, input original videos of subject X to the trained model, and convert videos where the faces are swapped to subject Y.
\item We create control groups. We cover black masks to face regions identical to the Faceswap operated face regions (masked videos). We blur the same face areas to create the second control group (blurred videos).
\item We run Openpose on the original, swapped, masked and blurred videos, and evaluate the keypoint location changes, taking Openpose result of original videos as ground truth.
\end{enumerate}
\subsection{Dataset}
We work on videos of patients with Parkinson's disease, a representative movement disorder. The experiment goes on UPDRS test videos of two subjects. Subject M is a male patient with UPDRS score 8, and subject F is a female patient with UPDRS score 31. Each subject's videos contain a series of tasks from one UPDRS trial, including talking, tremor, finger tapping, toe tapping, arising from chair, gait, and comprehensively record subject's movements. Subject M has 7176 frames extracted from videos, and subject F has 9282 frames.\\
\indent As for face-swapping target, we choose open-source videos of actor 01 (subject A) from the Deep Fake Detection Dataset\cite{googledata} provided by Google and JigSaw. A total of 3531 frames of subject A are selected for face swap model training. As for patients' training data, we select item 3.1 (speech) and item 3.2 (facial expression) frames from different trials to create a high quality training set for subject F (2320 frames) and subject M (1761 frames).
\subsection{Faceswap for Face-Swapping}

We choose Faceswap as the face changing tool. For every subject pairs to swap, a model is trained specifically. Here we use the original model consisting of a shared encoder and two decoders\cite{autoencoder}. The encoder encodes an input face image into a vector, which is a representation of the face, and the decoder recreates the face from the representation. Subject X and Y share the same encoder but each has its own decoder. During training, losses between input faces and recreated faces are calculated and weights of the encoder and decoders are updated. After training, an image of subject X could be swapped to subject Y after encoded by the shared encoder and recreated by subject Y's decoder. The whole structure is shown in Figure \ref{ae-train}. We train the original model with batch size 64. 
\subsection{Traditional De-Identification}
We adopt two traditional ways to de-identify subjects: black mask and blur. The black mask method covers a black rectangle box to face region. The blur method applies an average filter to face region, with kernel size half of the face detection bounding box. To assure manipulations are applied to the same face region, we record the face detection information of each swapped image from Faceswap log file and use the same region for masking and blurring.
\subsection{Openpose for Keypoint Detection}
Openpose is a widely used real-time multi-person system to jointly detect human keypoints developed by CMU Perceptual Computing Lab. Convolutional pose machine is used in Openpose, which consists of a sequence of convolutional networks that repeatedly produce 2D belief maps for the location of each part. At each stage, image features and the belief maps produced by the previous stage are used as input, with expanding receptive field. Prediction refines over stages.
\subsection{Face Recognition}
We use dlib's face recognition tool\cite{dlib} to quantify our face de-identification result. The dlib model\cite{dlib-face} is a ResNet network with 29 convolutional layers, a revised version of the ResNet-34 network designed for image recognition\cite{face-recognition}. It maps a face image to a 128-dimensional face descriptor. Descriptors of the same person are near to each other while descriptors from different people are far apart. By calculating the Euclidean distance of descriptors, it can be determined whether the input images are of the same person. Set the distance threshold to 0.6, the dlib model obtains an accuracy of 99.38\% on the standard LFW face recognition benchmark\cite{LFWTech,LFWTechUpdate}.
\subsection{Keypoint Evaluation}
We use keypoint evaluation metrics of Microsoft COCO dataset\cite{coco,ronchi} to quantify keypoints location changes. Object keypoint similarity (OKS) quantifies the similarity between predicted keypoints and ground truth keypoints:
$${\rm{OKS}} = \frac{{{\Sigma _i}\left[ {\exp \left( { - \frac{{d_i^2}}{{2{s^2}\kappa _i^2}}\delta \left( {{v_i} > 0} \right)} \right)} \right]}}{{{\Sigma _i}\left[ {\delta \left( {{v_i} > 0} \right)} \right]}}$$
The $d_i$'s are the Euclidean distances (errors) of detected keypoints and ground truth keypoints; the $v_i$'s are the visibility indices of ground truth keypoints ($>0$ if annotated); $s$ is the object scale; $\kappa_i$'s are per-keypoint constants that control falloff, determined by human annotator performance analysis. Each keypoint has a keypoint similarity that ranges between 0 and 1, and by averaging visible keypoint similarities over an object, we get the OKS. After setting a threshold of OKS value, the average precision (AP) and average recall (AR) are determined. In the COCO keypoint detection challenge, the average of AP at OKS=0.50:0.05:0.95 is the primary challenge metric. 
\begin{figure}[h]
\centering
\includegraphics[width=0.95\columnwidth]{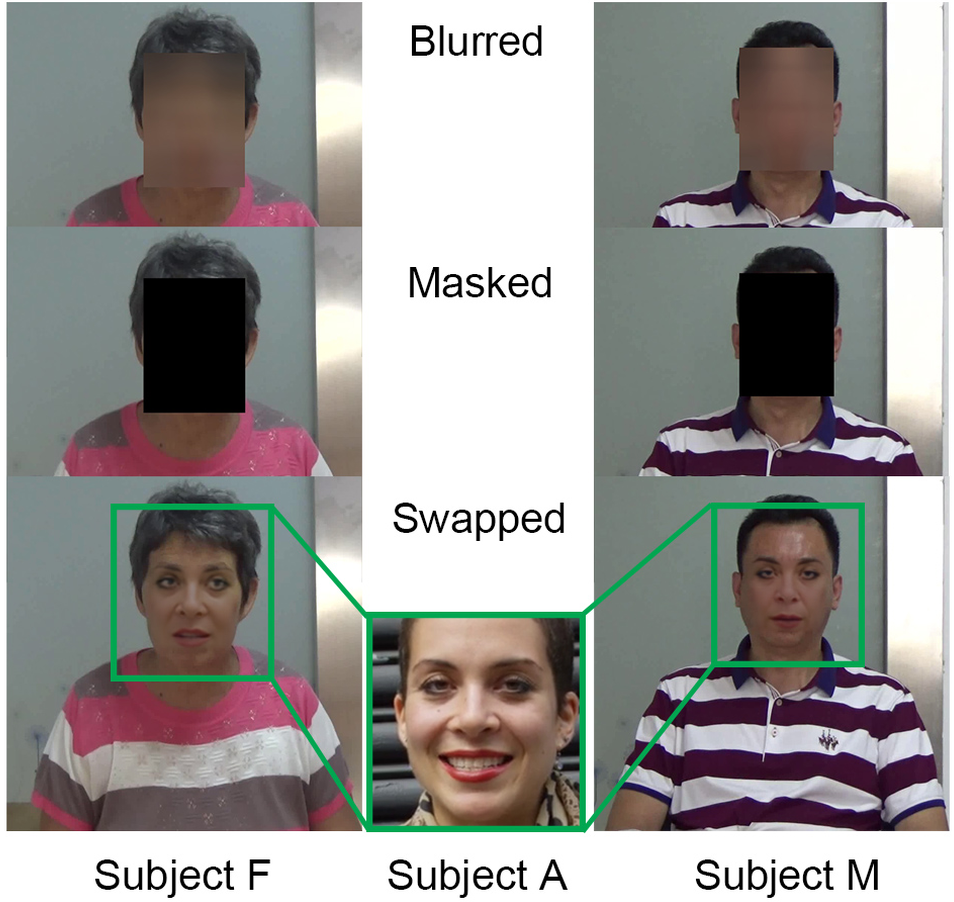} 
\caption{Images of subject F and M after de-identification and subject A (the target face).}
\label{face}
\end{figure}

\begin{table*}[t]
\centering
\caption{Face descriptor average distances and standard deviation. Given threshold $0.6$, the original faces are quite different between subjects ($>0.7$); the swapped faces are sufficiently similar to target subject A ($<0.5$); the swapped faces are sufficiently different from their corresponding original subjects ($>0.6$).}\smallskip
\begin{tabular}{lcccc}
\toprule
 			& Intra-subset  & To original  & To average original & To average subject A \\
\midrule
Swapped F 	& 0.190 (0.087) & 0.636 (0.033) & 0.629 (0.019) 	&  0.466 (0.027) \\
Swapped M 	& 0.176 (0.062) & 0.634 (0.029) & 0.631 (0.024) 	&  0.439 (0.029) \\ 
Original F 	& 0.181 (0.070) & 	-			& 	-				&  0.756 (0.022)\\ 
Original M 	& 0.136 (0.040) & -				 & -				 & 0.706 (0.022) \\ 
Original A 	& 0.276 (0.050) & -				 & -				 & - \\ 
\bottomrule
\end{tabular}
\label{dlibdist}
\end{table*}
\begin{figure}[t]
\centering
\includegraphics[width=0.9\columnwidth]{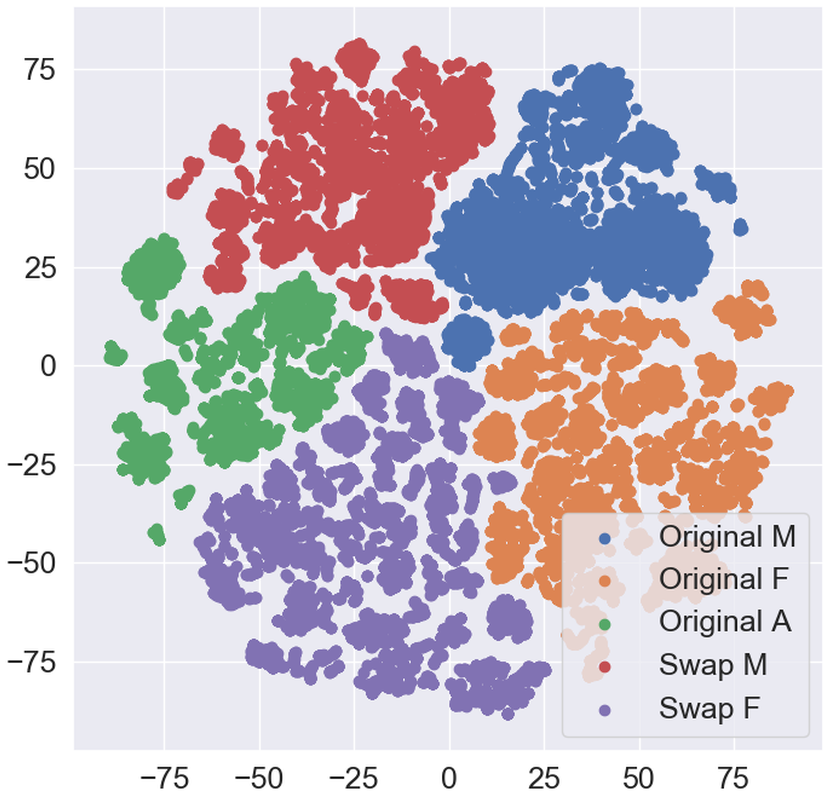} 
\caption{T-SNE cluster visualization of face descriptor of subjects' original and swapped images. Each point represents one face image. The points tend to cluster inside face subsets and segregate from other face subsets. This behavior implies that face swapping reliably de-identifies subjects.}
\label{tsne}
\end{figure}
\section{Result}
In this section, we present the results of employing our pipeline to UPDRS videos. We validate the reliability to make sure that the videos are unrecognizable after swapping. Then we evaluate the invariability of keypoints to confirm that face-swapping does not change keypoint information. Finally, we state the irreversibility of this de-identification method. Figure \ref{face} gives a sample of masked, blurred, and swapped images of subjects F and M\footnote{Demo videos:\newline \url{https://youtu.be/NZ2RbQxAq_w}\newline \url{https://youtu.be/1Lfok_vzEXM}}. Their original images are not shown for privacy concerns.\\
\indent During face-swapping, blurry images and images with no face detected are discarded. The output swapped set contains 6601 frames of subject F and 5129 frames of subject M. Masking and blurring are applied to the same images. Final image dataset contains 9 subsets: the swapped subject F/M, the masked subject F/M, the blurred subject F/M, and the original subset F/M/A. \\
\indent We run Faceswap and Openpose on a NVIDIA GTX TITAN X GPU. The Faceswap training takes 19 hours for the pairing of subject F and A, and 18 hours for subject M and A. The face-swapping of videos takes 120 minutes in total. Openpose keypoint detection takes 40 minutes in total.
\begin{figure}[t]
\centering
\includegraphics[width=0.9\columnwidth]{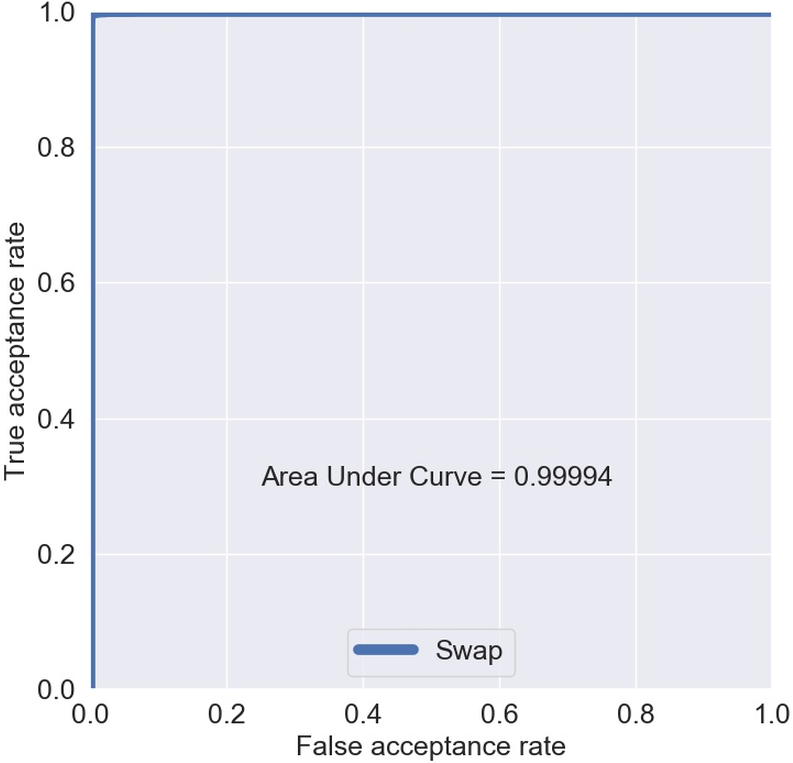} 
\caption{The ROC curve of swapped images. The curve shows that the swapped face is highly likely to be determined as target subject A, with little chance to be mistakenly determined as the original subject.}
\label{roc}
\end{figure}

\begin{figure}[h]
\centering
\includegraphics[width=1\columnwidth]{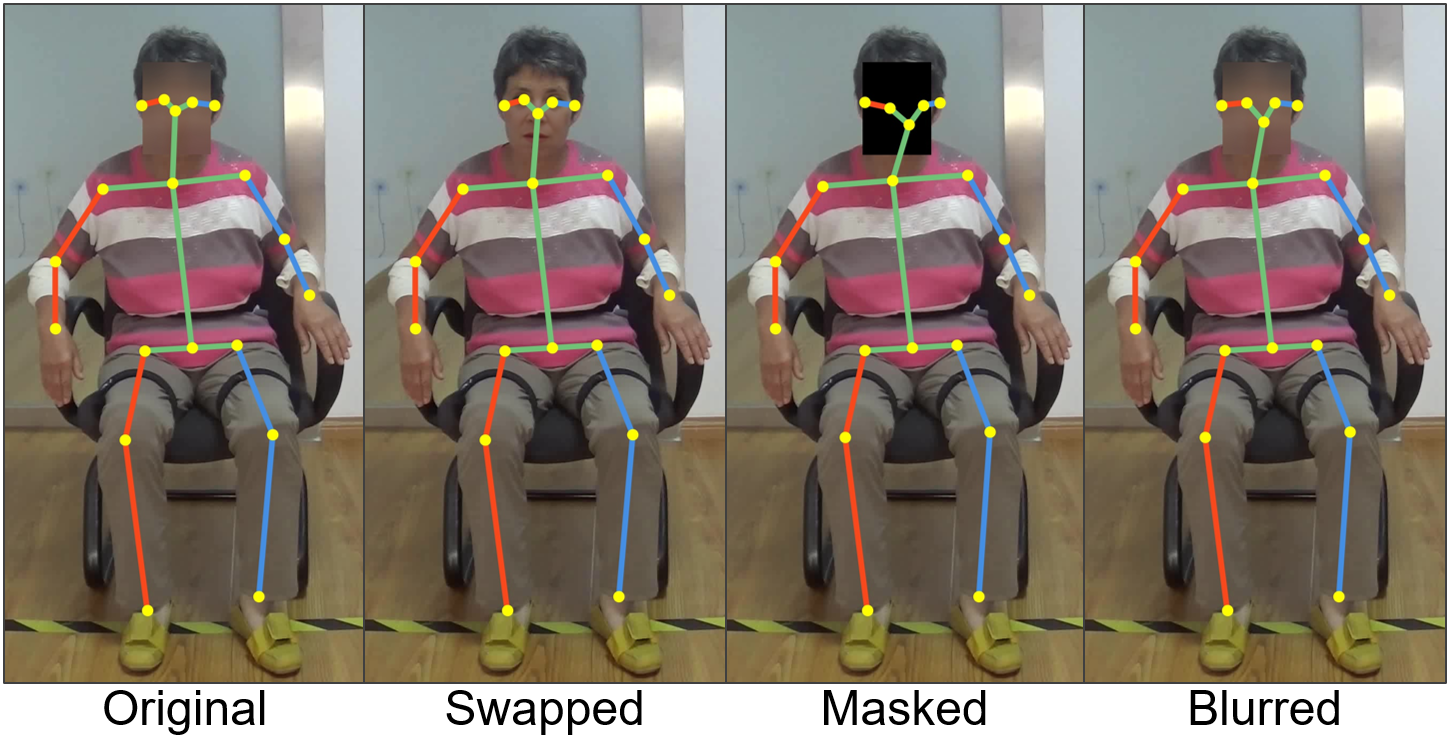} 
\caption{Sample keypoint detection result of an image from subject F during toe tapping task. Note that the original image's face is blurred after keypoint detection for privacy concerns.}
\label{kps-vis}
\end{figure}

\begin{figure}[h!]
\centering
\includegraphics[width=0.95\columnwidth]{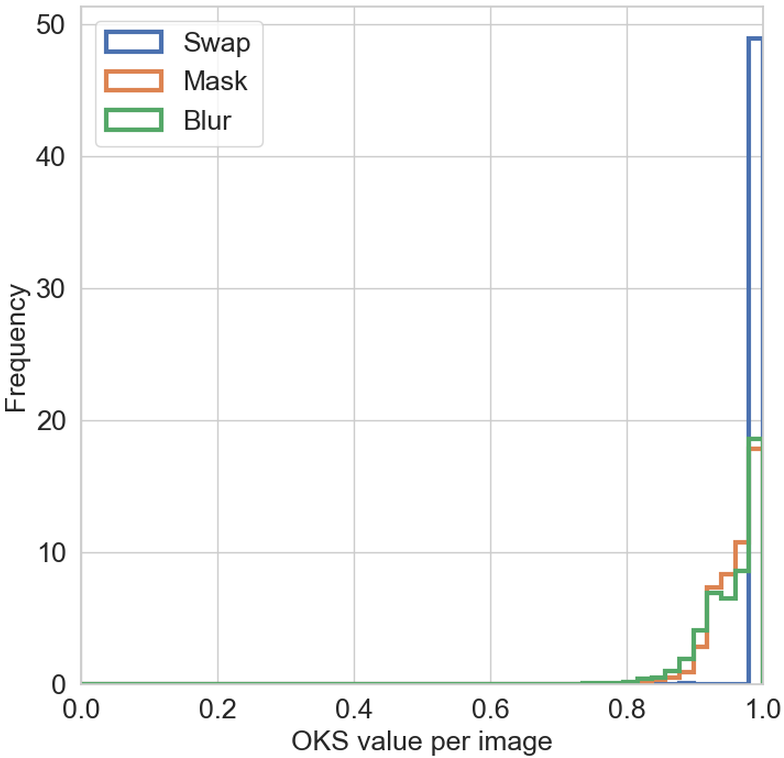} 
\caption{Image OKS distribution. The swapping method outperforms the other 2 methods. The main difference is at high OKS ($>$0.85) threshold.}
\label{oks-img}
\end{figure}

\begin{table}[h]
\centering
\caption{Performance on keypoint invariability. Swapped images show the highest AP and AR at any OKS threshold. Note that at extra high OKS threshold 0.95, AP and AR of swapped images are still close to 1.}\smallskip
\begin{tabular}{lccc}
\toprule
 				& Swapped  & Masked  & Blurred\\ 
\midrule
AP$^{0.5:0.95}$ & \textbf{0.993} & 0.933 & 0.916\\
AP$^{0.5}$ 		& \textbf{1.000} & \textbf{1.000} & \textbf{1.000}\\ 
AP$^{0.75}$ 	& \textbf{0.990} & \textbf{0.990} & \textbf{0.990}\\
AP$^{0.95}$ 	& \textbf{0.990} & 0.443 & 0.408\\ 
AR$^{0.5:0.95}$ & \textbf{1.000} & 0.960 & 0.948\\
AR$^{0.5}$ 		& \textbf{1.000} & \textbf{1.000} & \textbf{1.000}\\ 
AR$^{0.75}$ 	& \textbf{1.000} & \textbf{1.000} & 0.998\\ 
AR$^{0.95}$ 	& \textbf{0.998} & 0.654 & 0.606\\ 
\bottomrule
\end{tabular}
\label{eval}
\end{table}

\begin{figure*}[h!]
\centering
\includegraphics[width=0.95\textwidth]{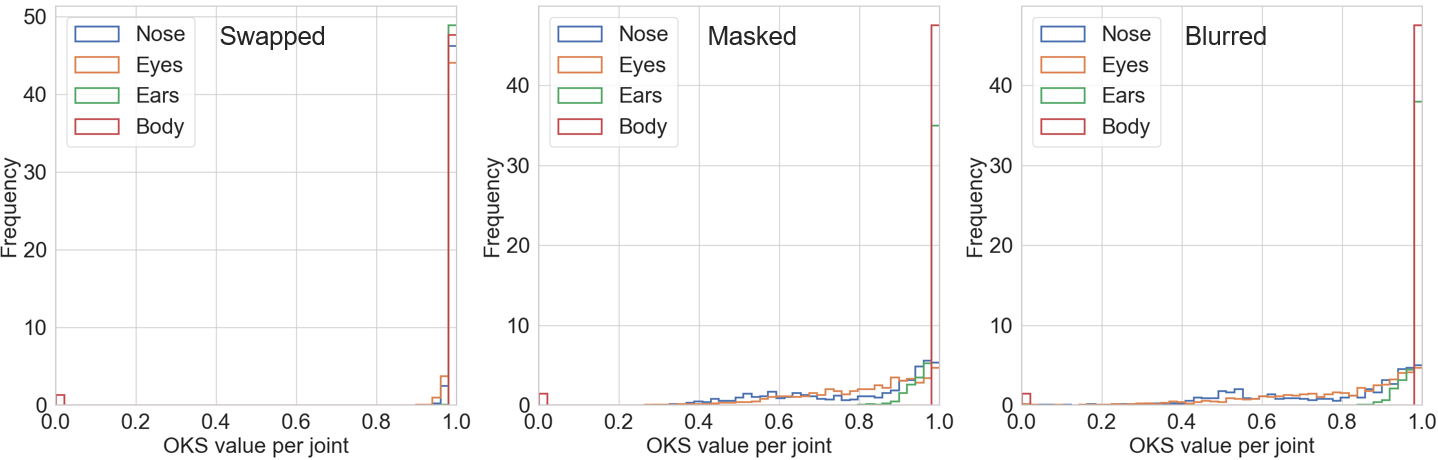} 
\caption{Single keypoint similarity distributions. It can be seen that body keypoints consistently show high keypoint similarities ($>$0.95, omitting few outliers), but head keypoints (nose, eyes and ears) behave differently between swapped output and traditional output. Masked and blurred data's head keypoints have dispersed distribution, and swapped data's head keypoints have concentrated high keypoint similarity distribution.}
\label{oks-swap}
\end{figure*}

\subsection{Reliability of Face De-Identification}
The swapped faces not only are intuitively distinct from original faces, but also meet the criteria that the Euclidean distances between face descriptors are greater than threshold $0.6$. Hence, the reliability of de-identification is verified.\\
\indent From Figure \ref{face}, intuitively the swapped faces are similar to the target. Figure \ref{tsne} gives the t-SNE\cite{tsne} result running on the dlib face descriptors of the swapped and original image subsets. The plot clearly shows that points (each point represents a face image) are close to each other if they come from the same subset. This cluster feature implies that the dlib face descriptor is able to distinguish face data here. In the figure, face subsets seem to have boundaries to separate them. And it further validates that the dlib method is applicable and implies that the face-swapping method de-identifies subjects. Notice that subject A's faces are far from subject F and M's original faces, while swapped faces are in the middle of original faces. This interesting distribution implies that face-swapping keeps some of the original features and generates new faces fitting to the original images. \\
\indent Table \ref{dlibdist} gives a quantified de-identification result that further confirms the reliability. The table lists the average Euclidean distances with standard deviation followed in parentheses. The \textit{Intra-subset} column calculates the distances to the subset's average descriptor. The \textit{To original} column calculates the distances between swapped and original descriptors. The \textit{To average original} calculates the distances between swapped descriptors and the corresponding average original descriptor. The \textit{To average subject A} calculates the distances between swapped descriptors and the average original descriptor of subject A. It is verified that the dlib face descriptors distinguish different people well in this work. All the subsets have a low intra-subset distance ($\ll 0.6$), while the average distance between original patients and subject A reaches the threshold 0.6. The face de-identification is successful, because both the \textit{To original} and the \textit{To average original} exceeds the threshold 0.6. Moreover, the generated faces are actually subject A, because their \textit{To average subject A} values are below the threshold 0.6. \\
\indent The receiver operating characteristic (ROC) curve (Figure \ref{roc}) gives a straight forward visualization, showing that the recreated faces are always to be identified as subject A under different threshold values. We define the true acceptance as the swapped image being sufficiently close to average subject A, and define the false acceptance as the swapped image being sufficiently close to average original subjects. From the curve we could see that nearly 100\% true acceptance rate is reached with a false acceptance rate close to 0. The area under curve is greater than 0.9999. The ROC curve shows that the swapped faces are always recognized as subject A and seldom recognized as the original subjects.
\subsection{Invariability of Keypoint Detection}
As mentioned above, de-identification methods should avoid changing keypoint detection outcomes. Our evaluation results show that Faceswap does the best job of keeping keypoints unchanged, and it successfully passes high precision keypoint detection tests. Figure \ref{kps-vis} displays images rendered with detected keypoints.\\
\indent Keypoints being totally invariant means that keypoint locations are identical. Smaller the change, better the de-identification method. In this case, our modifications are based on original images, so change calculation should be taken out between modified images and corresponding original images. We choose the COCO evaluation as our metric. The invariability determined by COCO evaluation shall be similar to other metrics. Here, we set Openpose keypoint detection of original images as ground truth, and compare the swapped, the masked, and the blurred results. The evaluation results are shown in table \ref{eval}. We could see that the swapped AP$^{0.5:0.95}$ (0.993) is significantly higher than masked AP$^{0.5:0.95}$ (0.933) and blurred AP$^{0.5:0.95}$ (0.916). The swapped AP at OKS threshold 0.95 (AP$^{0.95}$) remains 0.990, while the masked and blurred AP$^{0.95}$ drops drastically blow 0.45. Figure \ref{oks-img} shows the distribution of image OKS. The result shows that the swapped images' keypoint information is kept invariant while traditional methods change the keypoint information. \\
\indent To further investigate the exact changes made in different methods, the distribution of single keypoint similarity is shown in Figure \ref{oks-swap}. In swapped images, all keypoints are of high similarity, while in the masked and blurred images, keypoints from head have a dispersed distribution and consequently lower the overall OKS. The low head keypoint similarity of traditional methods is understandable: the face region is completely covered with black box or large kernel blur, so detection of head keypionts is more like a random guess. The face-swapping method recreates eyes, nose and ears and consequently leads to nearly unchanged keypoints. Body keypoints are less influenced by different methods. For body keypoints, the face-swapping method shows no significant improvements on keypoint invariability. The possible explanation of the body keypoint stability comes from Openpose's bottom-up feature. Openpose uses convolutional pose machine to predict keypoints, which has receptive fields of different sizes. Thus, modifications on the head area are unlikely to substantially change the prediction of body keypoints. Other keypoint detection tools with top-down feature might provide different result on body keypoints invariability.\\
\indent We confirm that compared to traditional methods, face-swapping does keep more movement information.
\subsection{Irreversibility of Face Swap}
The starting point of this work is to protect privacy. So it is a critical issue that whether original face identity could be reversely reconstructed from public accessible information. If there existed a way to attack the de-identified videos and recreate the original faces, privacy would be infringed.\\
\indent Consider this case: an attacker has a large amount of swapped videos of one subject and knows the structure of the deepfake model, but the attacker has no information about the subject's potential identity. The attacker intends to reconstruct the original face of the subject. To do that, the attacker must have the exact encoder and decoder of the subject pair. However, the model is trained locally by the video owner and not accessible to the attacker. In addition, it is impossible to learn about weights of a specific model with merely swapped videos where original faces are absent and only open-source faces present. Thus, it is hard to reversibly recreate the original faces as long as the trained model weights are unknown. In this way, privacy of subject is protected.
\section{Conclusion}
Deepfake technology has the ability to challenge human visual authenticity. It might create huge ethical and legal problems in the near future. In this paper, we provide a positive perspective on deepfake technology: it could help privacy protection and solve certain ethical concern, instead of being an ethical issue. Applying this method could promote medical video data sharing and improve medical research. It could advance human knowledge, for example, precise markerless movement tracking.\\
\indent In this work, we propose a pipeline for privacy protection that keeps original keypoint information. The core idea is to swap the faces so that the person is de-identified whilst the body keypoints and even the facial keypoints remain unchanged. We examine the reliability of privacy protection, the invariability of keypoints and the irreversibility of face change. The proposed pipeline has been shown to be effective for keypoint location preserving, robust to different people and recording conditions, and invulnerable to attacks.\\
\indent Further improvements are necessary to make this method more applicable. First, it relies heavily on face detection, and would be meaningless if face detection misses even one single face. This happens when the face is not big enough to be recognized by face detectors but still recognizable to humans, although we can solve it by manually screening missed frames for high-value medical videos. Next, in case of multi-person videos, the method might mess up with different people in the same image and generate monster faces. This method also performs worse on profiles, which impairs the continuity of facial information. In addition, hairstyle or clothing might accidentally release personal information. Further de-identification attempts on these parts would better protect privacy.\\
\indent To our best knowledge, this is the first work to show that deepfake technology is applicable to keypoint invariant de-identification, and to demonstrate that face-swapping approach could achieve privacy-preserving data sharing for high-value medical videos/images.
%
\begin{acks}
This work is sponsored by The National Key Research and Development Program of China (2016YFC0105502) and NSFC (81527901). All the authors are affiliated with the National Engineering Laboratory for Neuromodulation, School of Aerospace Engineering, Tsinghua University. Luming Li is also affiliated with: Precision Medicine \& Healthcare Research Center, Tsinghua-Berkeley Shenzhen Institute, Tsinghua University; IDG / McGovern Institute for Brain Research, Tsinghua University; Institute of Epilepsy, Beijing Institute for Brain Disorders. Correspondence to Yanan Sui and Luming Li.
\end{acks}

%
\bibliographystyle{ACM-Reference-Format}
\balance
\bibliography{zbq-reference.bib}


\begin{thebibliography}{27}


\ifx \showCODEN    \undefined \def \showCODEN     #1{\unskip}     \fi
\ifx \showDOI      \undefined \def \showDOI       #1{#1}\fi
\ifx \showISBNx    \undefined \def \showISBNx     #1{\unskip}     \fi
\ifx \showISBNxiii \undefined \def \showISBNxiii  #1{\unskip}     \fi
\ifx \showISSN     \undefined \def \showISSN      #1{\unskip}     \fi
\ifx \showLCCN     \undefined \def \showLCCN      #1{\unskip}     \fi
\ifx \shownote     \undefined \def \shownote      #1{#1}          \fi
\ifx \showarticletitle \undefined \def \showarticletitle #1{#1}   \fi
\ifx \showURL      \undefined \def \showURL       {\relax}        \fi
\providecommand\bibfield[2]{#2}
\providecommand\bibinfo[2]{#2}
\providecommand\natexlab[1]{#1}
\providecommand\showeprint[2][]{arXiv:#2}

\bibitem[\protect\citeauthoryear{Akhter and Black}{Akhter and Black}{2015}]%
        {mpi3d-dataset}
\bibfield{author}{\bibinfo{person}{Ijaz Akhter} {and}
  \bibinfo{person}{Michael~J Black}.} \bibinfo{year}{2015}\natexlab{}.
\newblock \showarticletitle{Pose-conditioned joint angle limits for 3D human
  pose reconstruction}. In \bibinfo{booktitle}{\emph{Proceedings of the IEEE
  conference on computer vision and pattern recognition}}.
  \bibinfo{pages}{1446--1455}.
\newblock


\bibitem[\protect\citeauthoryear{Barton and Hall}{Barton and Hall}{2015}]%
        {updrsvideo}
\bibfield{author}{\bibinfo{person}{Brandon~R Barton} {and}
  \bibinfo{person}{Deborah~A Hall}.} \bibinfo{year}{2015}\natexlab{}.
\newblock \bibinfo{booktitle}{\emph{Video Protocols and Techniques for Movement
  Disorders}}.
\newblock \bibinfo{publisher}{Oxford University Press}.
\newblock


\bibitem[\protect\citeauthoryear{Cao, Simon, Wei, and Sheikh}{Cao
  et~al\mbox{.}}{2017}]%
        {openpose-multi-person}
\bibfield{author}{\bibinfo{person}{Zhe Cao}, \bibinfo{person}{Tomas Simon},
  \bibinfo{person}{Shih-En Wei}, {and} \bibinfo{person}{Yaser Sheikh}.}
  \bibinfo{year}{2017}\natexlab{}.
\newblock \showarticletitle{Realtime multi-person 2d pose estimation using part
  affinity fields}. In \bibinfo{booktitle}{\emph{Proceedings of the IEEE
  Conference on Computer Vision and Pattern Recognition}}.
  \bibinfo{pages}{7291--7299}.
\newblock


\bibitem[\protect\citeauthoryear{Chesney and Citron}{Chesney and
  Citron}{2018}]%
        {fsbad2}
\bibfield{author}{\bibinfo{person}{Robert Chesney} {and}
  \bibinfo{person}{Danielle~Keats Citron}.} \bibinfo{year}{2018}\natexlab{}.
\newblock \showarticletitle{Deep fakes: a looming challenge for privacy,
  democracy, and national security}.
\newblock  (\bibinfo{year}{2018}).
\newblock


\bibitem[\protect\citeauthoryear{Deepfakes}{Deepfakes}{2017}]%
        {deepfakes}
\bibfield{author}{\bibinfo{person}{Deepfakes}.}
  \bibinfo{year}{2017}\natexlab{}.
\newblock \bibinfo{title}{Faceswap}.
\newblock \bibinfo{howpublished}{\url{https://github.com/deepfakes/faceswap}}.
\newblock
\newblock
\shownote{Accessed: 2019-09-10.}


\bibitem[\protect\citeauthoryear{Gafni, Wolf, and Taigman}{Gafni
  et~al\mbox{.}}{2019}]%
        {facebook}
\bibfield{author}{\bibinfo{person}{Oran Gafni}, \bibinfo{person}{Lior Wolf},
  {and} \bibinfo{person}{Yaniv Taigman}.} \bibinfo{year}{2019}\natexlab{}.
\newblock \showarticletitle{Live Face De-Identification in Video}. In
  \bibinfo{booktitle}{\emph{Proceedings of the IEEE International Conference on
  Computer Vision}}. \bibinfo{pages}{9378--9387}.
\newblock


\bibitem[\protect\citeauthoryear{Goetz, Tilley, Shaftman, Stebbins, Fahn,
  Martinez-Martin, Poewe, Sampaio, Stern, Dodel, et~al\mbox{.}}{Goetz
  et~al\mbox{.}}{2008}]%
        {updrs}
\bibfield{author}{\bibinfo{person}{Christopher~G Goetz},
  \bibinfo{person}{Barbara~C Tilley}, \bibinfo{person}{Stephanie~R Shaftman},
  \bibinfo{person}{Glenn~T Stebbins}, \bibinfo{person}{Stanley Fahn},
  \bibinfo{person}{Pablo Martinez-Martin}, \bibinfo{person}{Werner Poewe},
  \bibinfo{person}{Cristina Sampaio}, \bibinfo{person}{Matthew~B Stern},
  \bibinfo{person}{Richard Dodel}, {et~al\mbox{.}}}
  \bibinfo{year}{2008}\natexlab{}.
\newblock \showarticletitle{Movement Disorder Society-sponsored revision of the
  Unified Parkinson's Disease Rating Scale (MDS-UPDRS): scale presentation and
  clinimetric testing results}.
\newblock \bibinfo{journal}{\emph{Movement disorders: official journal of the
  Movement Disorder Society}} \bibinfo{volume}{23}, \bibinfo{number}{15}
  (\bibinfo{year}{2008}), \bibinfo{pages}{2129--2170}.
\newblock


\bibitem[\protect\citeauthoryear{Gu, Deligianni, Lo, Chen, and Yang}{Gu
  et~al\mbox{.}}{2018}]%
        {opgait}
\bibfield{author}{\bibinfo{person}{Xiao Gu}, \bibinfo{person}{Fani Deligianni},
  \bibinfo{person}{Benny Lo}, \bibinfo{person}{W Chen}, {and}
  \bibinfo{person}{Guang-Zhong Yang}.} \bibinfo{year}{2018}\natexlab{}.
\newblock \showarticletitle{Markerless gait analysis based on a single RGB
  camera}. In \bibinfo{booktitle}{\emph{2018 IEEE 15th International Conference
  on Wearable and Implantable Body Sensor Networks (BSN)}}. IEEE,
  \bibinfo{pages}{42--45}.
\newblock
\showISSN{2376-8894}
\urldef\tempurl%
\url{https://doi.org/10.1109/BSN.2018.8329654}
\showDOI{\tempurl}


\bibitem[\protect\citeauthoryear{Guo, He, Zhu, and Li}{Guo
  et~al\mbox{.}}{2018}]%
        {autoencoder}
\bibfield{author}{\bibinfo{person}{Yanzong Guo}, \bibinfo{person}{Wangpeng He},
  \bibinfo{person}{Juanjuan Zhu}, {and} \bibinfo{person}{Cheng Li}.}
  \bibinfo{year}{2018}\natexlab{}.
\newblock \showarticletitle{A Light Autoencoder Networks for Face Swapping}. In
  \bibinfo{booktitle}{\emph{Proceedings of the 2018 2nd International
  Conference on Computer Science and Artificial Intelligence}}. ACM,
  \bibinfo{pages}{459--462}.
\newblock


\bibitem[\protect\citeauthoryear{Harris}{Harris}{2018}]%
        {fsbad1}
\bibfield{author}{\bibinfo{person}{Douglas Harris}.}
  \bibinfo{year}{2018}\natexlab{}.
\newblock \showarticletitle{Deepfakes: False Pornography Is Here and the Law
  Cannot Protect You}.
\newblock \bibinfo{journal}{\emph{Duke Law \& Technology Review}}
  \bibinfo{volume}{17} (\bibinfo{year}{2018}), \bibinfo{pages}{99}.
\newblock


\bibitem[\protect\citeauthoryear{He, Zhang, Ren, and Sun}{He
  et~al\mbox{.}}{2016}]%
        {face-recognition}
\bibfield{author}{\bibinfo{person}{Kaiming He}, \bibinfo{person}{Xiangyu
  Zhang}, \bibinfo{person}{Shaoqing Ren}, {and} \bibinfo{person}{Jian Sun}.}
  \bibinfo{year}{2016}\natexlab{}.
\newblock \showarticletitle{Deep residual learning for image recognition}. In
  \bibinfo{booktitle}{\emph{Proceedings of the IEEE conference on computer
  vision and pattern recognition}}. \bibinfo{pages}{770--778}.
\newblock


\bibitem[\protect\citeauthoryear{Huang, Ramesh, Berg, and Learned-Miller}{Huang
  et~al\mbox{.}}{2007}]%
        {LFWTech}
\bibfield{author}{\bibinfo{person}{Gary~B. Huang}, \bibinfo{person}{Manu
  Ramesh}, \bibinfo{person}{Tamara Berg}, {and} \bibinfo{person}{Erik
  Learned-Miller}.} \bibinfo{year}{2007}\natexlab{}.
\newblock \bibinfo{booktitle}{\emph{Labeled Faces in the Wild: A Database for
  Studying Face Recognition in Unconstrained Environments}}.
\newblock \bibinfo{type}{{T}echnical {R}eport} 07-49.
  \bibinfo{institution}{University of Massachusetts, Amherst}.
\newblock


\bibitem[\protect\citeauthoryear{Jankovic}{Jankovic}{2008}]%
        {pd}
\bibfield{author}{\bibinfo{person}{Joseph Jankovic}.}
  \bibinfo{year}{2008}\natexlab{}.
\newblock \showarticletitle{Parkinson’s disease: clinical features and
  diagnosis}.
\newblock \bibinfo{journal}{\emph{Journal of neurology, neurosurgery \&
  psychiatry}} \bibinfo{volume}{79}, \bibinfo{number}{4}
  (\bibinfo{year}{2008}), \bibinfo{pages}{368--376}.
\newblock


\bibitem[\protect\citeauthoryear{King}{King}{2017}]%
        {dlib-face}
\bibfield{author}{\bibinfo{person}{Davis King}.}
  \bibinfo{year}{2017}\natexlab{}.
\newblock \bibinfo{title}{High Quality Face Recognition with Deep Metric
  Learning}.
\newblock
  \bibinfo{howpublished}{\url{http://blog.dlib.net/2017/02/high-quality-face-recognition-with-deep.html}}.
\newblock
\newblock
\shownote{Accessed 2019-10-20.}


\bibitem[\protect\citeauthoryear{King}{King}{2009}]%
        {dlib}
\bibfield{author}{\bibinfo{person}{Davis~E King}.}
  \bibinfo{year}{2009}\natexlab{}.
\newblock \showarticletitle{Dlib-ml: A machine learning toolkit}.
\newblock \bibinfo{journal}{\emph{Journal of Machine Learning Research}}
  \bibinfo{volume}{10}, \bibinfo{number}{Jul} (\bibinfo{year}{2009}),
  \bibinfo{pages}{1755--1758}.
\newblock


\bibitem[\protect\citeauthoryear{Learned-Miller}{Learned-Miller}{2014}]%
        {LFWTechUpdate}
\bibfield{author}{\bibinfo{person}{Gary B. Huang~Erik Learned-Miller}.}
  \bibinfo{year}{2014}\natexlab{}.
\newblock \bibinfo{booktitle}{\emph{Labeled Faces in the Wild: Updates and New
  Reporting Procedures}}.
\newblock \bibinfo{type}{{T}echnical {R}eport} UM-CS-2014-003.
  \bibinfo{institution}{University of Massachusetts, Amherst}.
\newblock


\bibitem[\protect\citeauthoryear{Lin, Maire, Belongie, Hays, Perona, Ramanan,
  Doll{\'a}r, and Zitnick}{Lin et~al\mbox{.}}{2014}]%
        {coco}
\bibfield{author}{\bibinfo{person}{Tsung-Yi Lin}, \bibinfo{person}{Michael
  Maire}, \bibinfo{person}{Serge Belongie}, \bibinfo{person}{James Hays},
  \bibinfo{person}{Pietro Perona}, \bibinfo{person}{Deva Ramanan},
  \bibinfo{person}{Piotr Doll{\'a}r}, {and} \bibinfo{person}{C~Lawrence
  Zitnick}.} \bibinfo{year}{2014}\natexlab{}.
\newblock \showarticletitle{Microsoft coco: Common objects in context}. In
  \bibinfo{booktitle}{\emph{European conference on computer vision}}. Springer,
  \bibinfo{pages}{740--755}.
\newblock


\bibitem[\protect\citeauthoryear{Lin, Wang, Lin, and Tang}{Lin
  et~al\mbox{.}}{2012}]%
        {deid1}
\bibfield{author}{\bibinfo{person}{Yuan Lin}, \bibinfo{person}{Shengjin Wang},
  \bibinfo{person}{Qian Lin}, {and} \bibinfo{person}{Feng Tang}.}
  \bibinfo{year}{2012}\natexlab{}.
\newblock \showarticletitle{Face swapping under large pose variations: A 3D
  model based approach}. In \bibinfo{booktitle}{\emph{2012 IEEE International
  Conference on Multimedia and Expo}}. IEEE, \bibinfo{pages}{333--338}.
\newblock


\bibitem[\protect\citeauthoryear{Maaten and Hinton}{Maaten and Hinton}{2008}]%
        {tsne}
\bibfield{author}{\bibinfo{person}{Laurens van~der Maaten} {and}
  \bibinfo{person}{Geoffrey Hinton}.} \bibinfo{year}{2008}\natexlab{}.
\newblock \showarticletitle{Visualizing data using t-SNE}.
\newblock \bibinfo{journal}{\emph{Journal of machine learning research}}
  \bibinfo{volume}{9}, \bibinfo{number}{Nov} (\bibinfo{year}{2008}),
  \bibinfo{pages}{2579--2605}.
\newblock


\bibitem[\protect\citeauthoryear{Mandery, Terlemez, Do, Vahrenkamp, and
  Asfour}{Mandery et~al\mbox{.}}{2015}]%
        {kit3d-dataset}
\bibfield{author}{\bibinfo{person}{Christian Mandery},
  \bibinfo{person}{{\"O}mer Terlemez}, \bibinfo{person}{Martin Do},
  \bibinfo{person}{Nikolaus Vahrenkamp}, {and} \bibinfo{person}{Tamim Asfour}.}
  \bibinfo{year}{2015}\natexlab{}.
\newblock \showarticletitle{The KIT whole-body human motion database}. In
  \bibinfo{booktitle}{\emph{2015 International Conference on Advanced Robotics
  (ICAR)}}. IEEE, \bibinfo{pages}{329--336}.
\newblock


\bibitem[\protect\citeauthoryear{Mosaddegh, Simon, and Jurie}{Mosaddegh
  et~al\mbox{.}}{2014}]%
        {deid2}
\bibfield{author}{\bibinfo{person}{Saleh Mosaddegh}, \bibinfo{person}{Loic
  Simon}, {and} \bibinfo{person}{Fr{\'e}d{\'e}ric Jurie}.}
  \bibinfo{year}{2014}\natexlab{}.
\newblock \showarticletitle{Photorealistic face de-identification by
  aggregating donors’ face components}. In \bibinfo{booktitle}{\emph{Asian
  Conference on Computer Vision}}. Springer, \bibinfo{pages}{159--174}.
\newblock


\bibitem[\protect\citeauthoryear{Rossler, Cozzolino, Verdoliva, Riess, Thies,
  and Niessner}{Rossler et~al\mbox{.}}{2019}]%
        {googledata}
\bibfield{author}{\bibinfo{person}{Andreas Rossler}, \bibinfo{person}{Davide
  Cozzolino}, \bibinfo{person}{Luisa Verdoliva}, \bibinfo{person}{Christian
  Riess}, \bibinfo{person}{Justus Thies}, {and} \bibinfo{person}{Matthias
  Niessner}.} \bibinfo{year}{2019}\natexlab{}.
\newblock \showarticletitle{FaceForensics++: Learning to Detect Manipulated
  Facial Images}. In \bibinfo{booktitle}{\emph{The IEEE International
  Conference on Computer Vision (ICCV)}}.
\newblock


\bibitem[\protect\citeauthoryear{Ruggero~Ronchi and Perona}{Ruggero~Ronchi and
  Perona}{2017}]%
        {ronchi}
\bibfield{author}{\bibinfo{person}{Matteo Ruggero~Ronchi} {and}
  \bibinfo{person}{Pietro Perona}.} \bibinfo{year}{2017}\natexlab{}.
\newblock \showarticletitle{Benchmarking and error diagnosis in multi-instance
  pose estimation}. In \bibinfo{booktitle}{\emph{Proceedings of the IEEE
  International Conference on Computer Vision}}. \bibinfo{pages}{369--378}.
\newblock


\bibitem[\protect\citeauthoryear{Silbey and Hartzog}{Silbey and
  Hartzog}{2018}]%
        {fsgood}
\bibfield{author}{\bibinfo{person}{Jessica Silbey} {and}
  \bibinfo{person}{Woodrow Hartzog}.} \bibinfo{year}{2018}\natexlab{}.
\newblock \showarticletitle{The Upside of Deep Fakes}.
\newblock \bibinfo{journal}{\emph{Maryland Law Review}}  \bibinfo{volume}{78}
  (\bibinfo{year}{2018}), \bibinfo{pages}{960}.
\newblock


\bibitem[\protect\citeauthoryear{Simon, Joo, Matthews, and Sheikh}{Simon
  et~al\mbox{.}}{2017}]%
        {openpose-hand}
\bibfield{author}{\bibinfo{person}{Tomas Simon}, \bibinfo{person}{Hanbyul Joo},
  \bibinfo{person}{Iain Matthews}, {and} \bibinfo{person}{Yaser Sheikh}.}
  \bibinfo{year}{2017}\natexlab{}.
\newblock \showarticletitle{Hand keypoint detection in single images using
  multiview bootstrapping}. In \bibinfo{booktitle}{\emph{Proceedings of the
  IEEE conference on Computer Vision and Pattern Recognition}}.
  \bibinfo{pages}{1145--1153}.
\newblock


\bibitem[\protect\citeauthoryear{Wei, Ramakrishna, Kanade, and Sheikh}{Wei
  et~al\mbox{.}}{2016}]%
        {openpose-cpm}
\bibfield{author}{\bibinfo{person}{Shih-En Wei}, \bibinfo{person}{Varun
  Ramakrishna}, \bibinfo{person}{Takeo Kanade}, {and} \bibinfo{person}{Yaser
  Sheikh}.} \bibinfo{year}{2016}\natexlab{}.
\newblock \showarticletitle{Convolutional pose machines}. In
  \bibinfo{booktitle}{\emph{Proceedings of the IEEE Conference on Computer
  Vision and Pattern Recognition}}. \bibinfo{pages}{4724--4732}.
\newblock


\bibitem[\protect\citeauthoryear{Xue, Sayana, Darke, Shen, Hsieh, Luo, Li,
  Downing, Milstein, and Fei-Fei}{Xue et~al\mbox{.}}{2018}]%
        {opgait2}
\bibfield{author}{\bibinfo{person}{David Xue}, \bibinfo{person}{Anin Sayana},
  \bibinfo{person}{Evan Darke}, \bibinfo{person}{Kelly Shen},
  \bibinfo{person}{Jun-Ting Hsieh}, \bibinfo{person}{Zelun Luo},
  \bibinfo{person}{Li-Jia Li}, \bibinfo{person}{N~Lance Downing},
  \bibinfo{person}{Arnold Milstein}, {and} \bibinfo{person}{Li Fei-Fei}.}
  \bibinfo{year}{2018}\natexlab{}.
\newblock \showarticletitle{Vision-Based Gait Analysis for Senior Care}.
\newblock \bibinfo{journal}{\emph{arXiv preprint arXiv:1812.00169}}
  (\bibinfo{year}{2018}).
\newblock


\end{thebibliography}

\end{document}